\documentclass[a4paper,fleqn]{cas-dc}

\usepackage[numbers]{natbib}
\usepackage{multirow}
\usepackage{tabularx}
\usepackage{graphicx}
\usepackage{float}
\usepackage{subfigure}
\usepackage{threeparttable}
\usepackage{ulem}
\def\tsc#1{\csdef{#1}{\textsc{\lowercase{#1}}\xspace}}
\tsc{WGM}
\tsc{QE}
\tsc{EP}
\tsc{PMS}
\tsc{BEC}
\tsc{DE}

\begin{document}
\let\WriteBookmarks\relax
\def\floatpagepagefraction{1}
\def\textpagefraction{.001}

\title [mode = title]{3DTMDet: A Dual-Path Synergy Network of Transformer and SSM for 3D Object Detection in Point Clouds}                      

\author[1]{Bingwen Qiu}[style=chinese]
\author[1]{Yuan Liu}[style=chinese]     %
\author[2]{Junqi Bai}[style=chinese]
\author[1]{Tong Jiang}[style=chinese]
\author[1]{Ben Liang}[style=chinese]
\author[3]{Fangzhou Chen}[style=chinese]
\author[1,5]{Xiubao Sui}[style=chinese, orcid=0000-0003-0464-464X]
\cormark[1]
\author[1,4]{Qian Chen}[style=chinese]

\affiliation[1]{organization={School of Electronic and Optical Engineering},
            addressline={Nanjing University of Science and Technology}, 
            city={Nanjing},
            postcode={210094}, 
            state={Jiangsu},
            country={China}
                }

\affiliation[2]{organization={The 28th Research Institute of China Electronics Technology Group Corporation},
            city={Nanjing},
            postcode={210007}, 
            state={Jiangsu},
            country={China}
                }
\affiliation[3]{organization={College of Astronautics},
            addressline={Nanjing University of Aeronautics and Astronautics}, 
            city={Nanjing},
            postcode={210016}, 
            state={Jiangsu},
            country={China}
                }
\affiliation[4]{organization={School of Information and Communication Engineering},
            addressline={North University of China}, 
            city={Taiyuan},
            postcode={030051}, 
            state={Shanxi},
            country={China}
                }
\affiliation[5]{organization={State key Laboratory of Extreme Environment Optoelectronic Dynamic Measurement Technology and Instrument},
            addressline={Nanjing University of Science and Technology}, 
            city={Nanjing},
            postcode={210094}, 
            state={Jiangsu},
            country={China}
                }
                
\cortext[cor1]{Corresponding author at: School of Electronic and Optical Engineering, Nanjing University of Science and Technology, Nanjing 210094, China\\
E-mail address: sxbhandsome@163.com (X. Sui).}

\begin{abstract}
A fundamental challenge in point cloud object detection lies in the conflict between the extreme sparsity of distant points and the need for remote context understanding. The existing methods typically use 1D serialization to expand the receptive field, which inevitably discards already scarce local geometric details and reduces detection of distant and small objects. To address this issue, we propose 3DTMDet, a novel detection network that synergistically combines state space models (Mamba) with Transformers. The core idea is to utilize SSM's linear complexity and advantages in long sequence modeling to effectively capture global interactions between sparse and distant points, while using Transformer modules with local attention to encode fine-grained geometric structures in local point sets, preserving accurate shape information. We propose the 3D Hybrid Mamba Transformer (3DHMT) block, which uses an SSM-Attention-SSM pipeline to balance global context understanding and local detail preservation, effectively alleviating the tension between receptive field enlargement and geometric preservation in remote detection. In addition, we introduced a voxel generation block inspired by LiDAR physics, which diffuses features along the sensor observation direction to reconstruct the complete object structure of occlusion and distant areas. Extensive experiments conducted on the KITTI and ONCE datasets have shown that 3DTMDet outperforms state-of-the-art detectors. The code is available at https://github.com/QiuBingwen/3DTMDet.
\end{abstract}


\maketitle

\section{INTRODUCTION}

Point cloud object detection serves as a fundamental perception task in autonomous driving and robotics\cite{Alaba2022ASO, Wu2024ASO}, where accurately identifying objects at long ranges is critical for system safety. However, LiDAR sensors inherently suffer from a quadratic decay in point density with distance, causing distant objects to be represented by only a handful of extremely sparse measurements\cite{WenwuChen2024DeeplearningenabledTS, QianEnsemble}. This severe sparsity leads to a collapse of local geometric structures, making fine-grained shape modeling, part-level recognition, and precise pose estimation nearly impossible. Simultaneously, long-range scenes involve more complex inter-object interactions—such as occlusions, truncations, and cluttered backgrounds—which demand a large receptive field to capture long-range contextual dependencies. Yet, expanding the receptive field typically relies on progressive downsampling and feature abstraction, which further dilutes the already scarce local details of distant targets. Consequently, a fundamental dilemma arises: the need to jointly achieve long-range global interaction understanding and near-field local fine-grained modeling in an extremely sparse point regime, where the two objectives conflict with each other\cite{EBLG, SVP}.

Faced with this dilemma, existing methods have exposed various limitations. Point based methods\cite{PointRCNN, 3DSSD, PDMSSD}, use farthest point sampling and neighborhood aggregation based on ball queries to perceive structure. However, on sparse distant objects, sampling becomes inefficient and expensive, and random downsampling can easily discard a few foreground points, resulting in frequent missed small targets or occluded targets. Voxel based methods, such as voxel networks, point columns, and center points, quantize point clouds into regular grids, introduce quantization errors, and smooth the fine geometric shapes of small distant objects. However, the average pooling encoding within each voxel further sacrifices the point by point geometric accuracy. In order to capture long-distance interactions, these methods stack downsampling layers, which exacerbate voxel sparsity and gradually eliminate local features that have already blurred distant targets. Transformer based detectors\cite{ASCFormer, M3DETR, SFNet, GroupFree3D, KPTr} utilize self attention mechanisms for modeling, but their quadratic complexity has poor scalability in large outdoor point clouds. To alleviate this situation, window based methods\cite{DSVT} limit attention to local regions, but this often impairs the model's ability to understand the overall scene structure. The methods based on the linear framework\cite{PointMamba, Lu2025ExploringTS, HyMamba, MambaOcc, LION} serializing 3D voxels into 1D sequence, sacrifice spatial continuity and damage fine-grained local representation. When point clouds are too sparse, they exhibit geometric ambiguity, inevitably damaging the already scarce geometric information of distant instances. We believe that an optimal 3D encoder can complement the tasks of efficient global feature interaction and accurate local feature extraction.

Another persistent challenge in LiDAR-based detection is the varying density of point clouds\cite{Jung2025AnalysisOO, Liu2025AUV, WANG2023112840}. Objects distant from the sensor often appear highly sparse, consisting of only a few points, which degrades feature representation and detection accuracy. Standard voxel generation methods fail to account for the physical imaging characteristics of LiDAR, leaving empty voxels in regions that are likely occupied but occluded or unsampled, thereby fragmenting object shapes.

To address these limitations, we propose 3DTMDet, a novel hybrid 3D object detection network that unifies the long-range efficiency of SSM and the high-fidelity geometric modeling capability of Attention in a principled synergy pipeline, rather than simple module stacking. Specifically, we design a bidirectional Serialized-Mamba block to capture global dependencies with linear complexity while alleviating spatial structure damage caused by conventional one-dimensional serialization. Meanwhile, we introduce a cross-attention Grouped-Transformer Block with dual Hilbert ordering and KNN-based relative position encoding to compensate for fine-grained local geometric details lost during serialization. Furthermore, to tackle the severe sparsity of distant LiDAR points, we propose a physics-inspired Voxel Generation block that diffuses features along the sensor viewing direction to reconstruct complete object structures in sparse and occluded regions. By integrating these designs into a repeated 3D Hybrid Mamba Transformer (3DHMT) building block, our framework achieves robust global–local complementary feature learning, effectively resolving the conflicts between efficiency and precision in existing 3D representation learning paradigms.


In summary, our contributions are as follows:
\begin{itemize}
\item We propose a hybrid 3D backbone that synergizes SSM’s global linear scaling with the Transformer’s local geometric precision. This ensures global scene comprehension without sacrificing the fine-grained structural integrity of small objects.

\item We implement a window-based Attention module utilizing Hilbert-curve grouping and relative position encoding. This serves as a spatial-rectifier that re-establishes local 3D dependencies, providing a mathematically grounded approach to processing non-Euclidean point cloud data in a sequence-based framework.

\item We introduce a Voxel Generated Block (VGB) that moves beyond passive feature extraction. By encoding the geometric priors of LiDAR imaging (the "occlusion-occupation" principle), the VGB performs active feature diffusion to restore fragmented object shapes in sparse, distant regions.
\end{itemize}

\section{RELATED WORK}

\subsection{Transformers in 3D Object Detection}
With its attention mechanism, Transformer is highly adept at flexibly modeling relationships between features and has driven significant advancements in 3D object detection. Mao et al. \cite{VoTr} first proposed local attention and expanded attention, and designed voxel query, so that the attention mechanism can be implemented on sparse voxels. Wang  et al. \cite{DSVT} proposes Dynamic Sparse Window Attention, which divides a series of local regions based on the sparsity of each window and then computes the features of all regions in a fully parallel manner. Liang  et al.\cite{DRF-SSD} proposes a Hybrid transformer module that combines the vote net with the transformer, utilizing cross-attention to increase focus on global features while retaining detailed features. Cai et al.\cite{DSTR} proposes a novel dual-scene transformer pipeline (DSTR), which includes a global scene integration (GSI) module, a local scene integration (LSI) module, and a dual-scene fusion (DSF) module, aiming to address the deficiencies in global scene fusion. Zhou et al.\cite{Pillar-CT3D++} proposes a self-attention feature encoding module that generates point cloud feature representations with higher information density, enabling better understanding of the spatial relationships and contextual information of objects in the point cloud scene. Xie et al.\cite{Mine-SSD} introduced a dual-head self-correlation module that dynamically adjusts the aggregation radius of candidate points through a radius-adaptive grouping mechanism, capturing key features of non-conventional objects to support multi-scale feature processing. Wan et al.\cite{Bridging} divided image features into multiple regions and employed region-to-pillar cross-attention to convey semantic information from the image to the corresponding LIDAR pillar.

Despite these successes, the quadratic complexity introduced by the core mechanism of Transformer, namely self-attention, remains a significant factor limiting its effectiveness. As the number of voxels in each group increases, algorithm complexity and resource consumption grow quadratically, indicating that large-scale grouping will lead to unacceptable resource consumption. To circumvent this issue, most Transformer-based 3D detectors limit the size of attention mechanism groups. Unfortunately, this artificial spatial partitioning severely restricts the receptive field, inevitably weakening the Transformer's ability to model long-range dependencies. Therefore, algorithms based purely on Transformer struggle to comprehend the overall global context and spatial distribution of the entire 3D scene.

In this work, we propose a window-based Transformer block, which utilize the cross-attention and the relative position encoding for local feature extraction. This block focuses solely on extracting local detail information, while the task of remote feature interaction is handled by the Mamba module.

\begin{figure}
	\centering
	\includegraphics[width=0.47\textwidth]{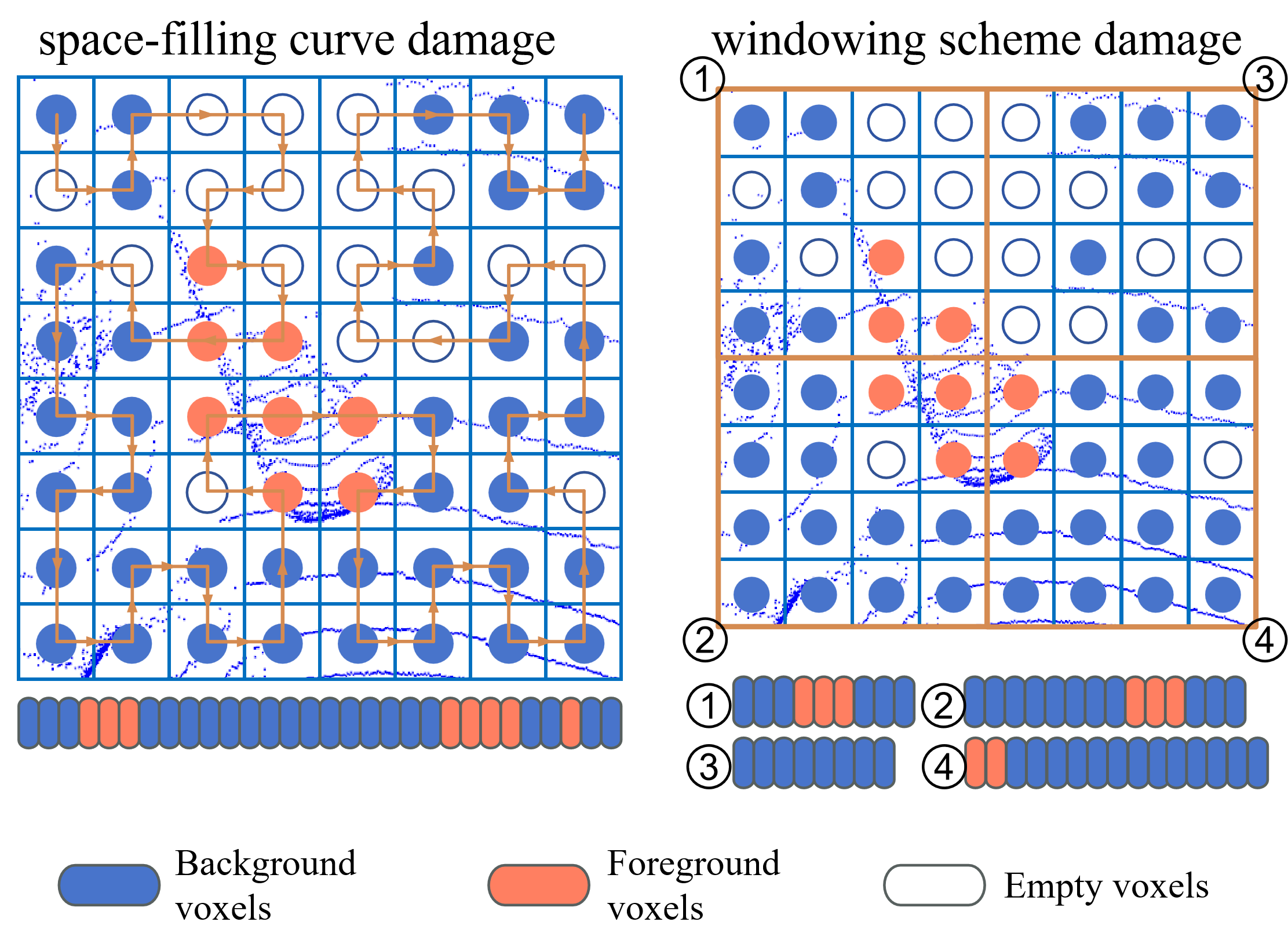}
	\caption{Illustration of the damage to the geometric structure  caused by space-filling curves and windowing schemes. Both of these schemes may lead to the division of the unified target into different positions in the sequence, or even different sequences. This can lead to potential instability in linear operators.}
	\label{structure_dest}
\end{figure}

\subsection{Mamba in 3D Tasks}
Mamba\cite{Mamba} has adapted in computer vision tasks due to its efficiency\cite{LIU2025127905}. Zhang et al. \cite{VoxelMamba} adopts a non-clustering strategy to serialize the entire voxel space into a single sequence, and proposes the establishment of a hierarchical structure using dual-scale SSM blocks, enabling the realization of a larger receptive field in the one-dimensional serialized curve. Liu et al.\cite{LION} proposed a simple and effective window framework based on Linear Group RNN for accurate 3D object detection. Ning et al.\cite{MambaFusion} utilized Hybrid Mamba Block for local and global context learning, and for the first time demonstrates that a pure Mamba module can achieve efficient dense global fusion while ensuring top-notch performance in camera-LiDAR multi-modal 3D object detection. Li et al.\cite{3DET-Mamba} introduced a novel Dual Mamba module modeling point clouds from the perspectives of spatial distribution and continuity.

The Mamba based algorithm has shown significant advantages in long-distance interactions. However, Mamba processes data in order, so 3D point clouds or voxels must be flattened into 1D sequences, which severely destroys the geometric structure of the data. In order to alleviate the problems caused by flattening operations, existing methods commonly use complex space filling curves. However, this forced 1D serialization still inevitably destroys the spatial continuity of the original scene and seriously damages the fine local geometric structure of the object.

Therefore, pure Mamba architecture typically produces smooth representations, lacking the high fidelity local details required for accurate detection of small, sparse, or geometrically complex objects such as pedestrians and cyclists. The existing methods have proposed various solutions to compensate for the damage caused by Mamba algorithm to three-dimensional geometric structures, such as window opening schemes and space filling curve combinations, but still cannot avoid the damage to geometric structures, as shown in Figure \ref{structure_dest}.

In this work, we propose a non-window Mamba block, which serialize voxels in an extremely simple order for long-range feature interaction.This module focuses on remote feature interaction without using any windowing or spatial padding schemes, making it simple and easy to operate. Its lack of geometric detail modeling capability is compensated by the transformer module.

\subsection{Hybrid Structures for Transformer and Mamba}

\begin{figure*}[htbp]
    \centering
    \includegraphics[width=1\textwidth, keepaspectratio]{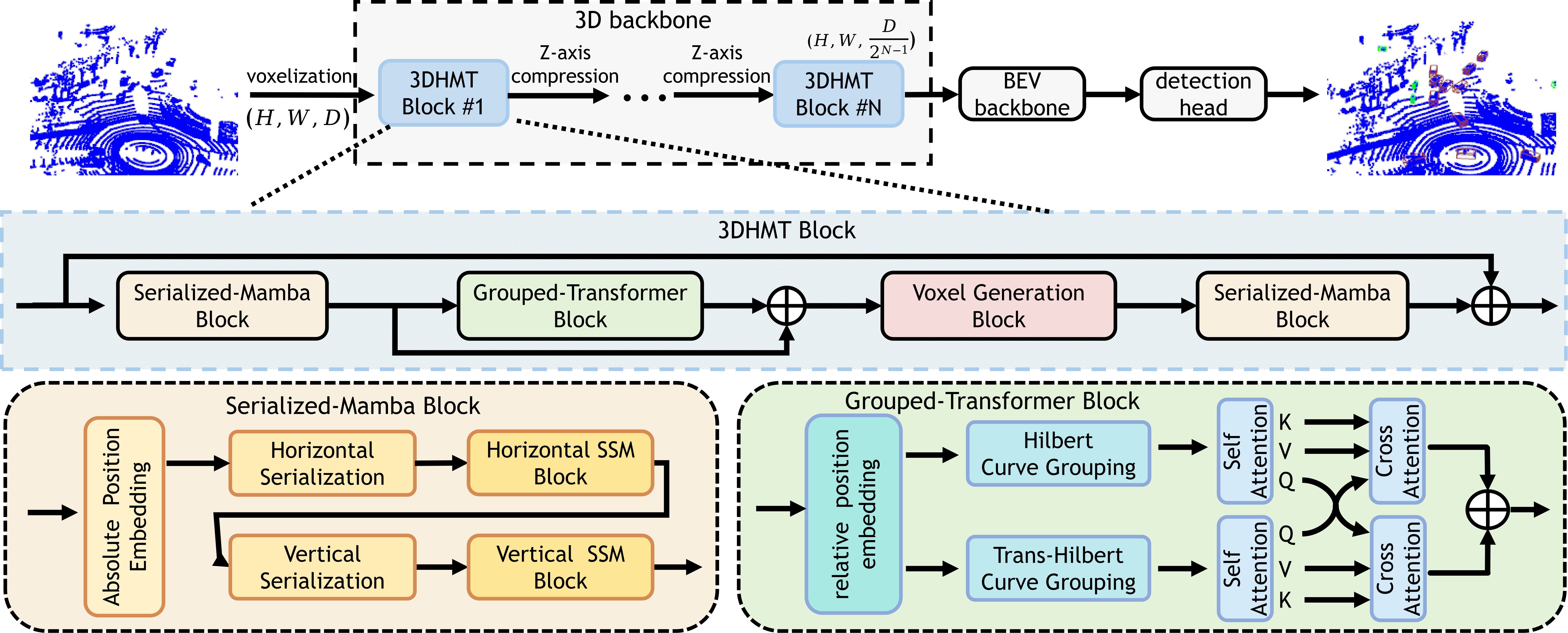}
    \caption{\textbf{Top:} The overall architecture of our proposed 3DTMDet. First, the point cloud data is converted into voxels and fed into a 3D backbone. The 3D backbone consists of multiple 3D Hybrid Mamba Transformer (3DHMT) Blocks, with voxels being compressed along Z-axis between modules. Internally, the 3DHMT Block consists of two Serialized-Mamba Blocks, one Grouped-Transformer Block and one Voxel Generation Block. The Serialized-Mamba Blocks and Grouped-Transformer Block are responsible for long-distance feature interaction and local feature extraction, respectively. The Voxel Generation Blocks are applied to enhance the sparse feature of objects from the perspective of sensors. Then, the voxels are converted into dense BEV and extract BEV features. Finally, it is fed into the dense detection head to produce detection results.
    \textbf{Bottom:} The illustrations of the Serialized-Mamba Block and the Grouped-Transformer Block. The symbol $\oplus$ denotes the add operation.}
    \label{figurelabel_overall}
    
\end{figure*}

Recognizing the bottlenecks of Transformers and the geometric limitations of Mamba, recent research has begun exploring their complementary abilities. The synergy is highly intuitive: Mamba can construct global, long-distance context efficiently over massive token sequences with linear complexity, while Transformers can be deployed strictly within small local windows to preserve and extract high-fidelity geometric details without blowing up resource consumption.

The method combining Mamba with Transformer has also caught the attention of researchers. Inaganti et al.\cite{MambaTron} proposed MambaTron, which is a Mamba-Transformer block, capable of achieving single-modality and cross-modality reconstruction, including view-guided point cloud completion. This is the earliest attempt to implement a Mamba-based Cross-attention analogy. Li et al.\cite{PoinTramba} introduces PoinTramba, a groundbreaking hybrid framework that synergizes the analytical capabilities of Transformer with the excellent computational efficiency of Mamba to enhance point cloud analysis. Wang  et al.\cite{HybridTM} proposed an inner-layer hybrid strategy that combines attention with Mamba at a finer granularity, enabling the simultaneous capture of long-range dependencies and fine-grained local features in 3D semantic segmentation.

Although the above attempts have been made, the existing hybrid frameworks still do not address the core pain points in 3D voxel detection: spatial damage in Mamba and complexity bottleneck in Transformer.

Building upon these concepts, this work proposes a  hybrid backbone that explicitly solves spatial damage and complexity bottlenecks via a synergistic global–local pipeline, making principled contributions to 3D neural architecture design.

\section{METHODOLOGY}

\subsection{Overview of the Method}

We propose 3DTMDet, a hybrid voxel-based 3D detector built on a novel dual-path synergy backbone. The pipeline follows standard voxelization, 3D backbone encoding, BEV feature conversion, and detection head, presented in Figure\ref{figurelabel_overall}. Our key innovation lies in the 3D backbone composed of stacked 3DHMT blocks.

Each 3DHMT block follows a SSM–Attention–SSM design:
\begin{itemize}
\item The first Serialized-Mamba Block performs bidirectional horizontal/vertical serialization to model long-range global dependencies while preserving spatial continuity.
\item The Grouped-Attention Block extracts high-fidelity local geometric features via cross-attention and relative position encoding, compensating structural loss from SSM serialization.
\item The second Serialized-Mamba Block further propagates enhanced local features across the global space. Simultaneously utilizing MAMBA's powerful auto-regressive ability to repair potential errors caused by Generated Blocks
\item Voxel Generation Block enhances sparsity by generating pseudo voxels along LiDAR viewing direction.
\end{itemize}

This global–local complementary pipeline is not a simple module combination, but a synergistic design that resolves the inherent weaknesses of pure SSM and pure Transformer.

\subsection{Serialized-Mamba Block}


\begin{figure}
	\centering
	\includegraphics[width=0.48\textwidth]{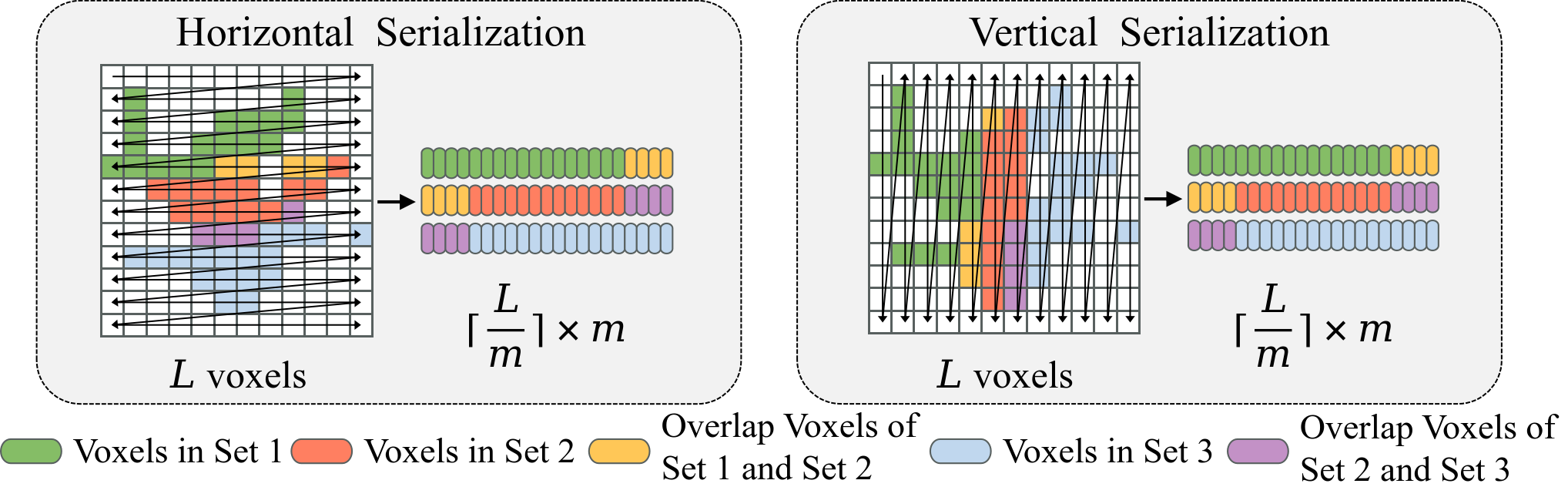}
	\caption{Illustration of the serialization scheme in Serialized-Mamba Block. All voxels are sorted into equally sized groups along the X or Y axis, and arranged from low to high along the Z-axis. Each adjacent sets share the equal-sized overlapping voxels. }
	\label{figurelabel_Mamba_serialization}
\end{figure}

To apply the Mamba module to the voxel data, we serialize the sparsely distributed voxels. First, the point cloud is converted into a total of $L$ voxels. As shown in figure \ref{figurelabel_Mamba_serialization}, these voxels are sorted along the $X$ and $Y$ axes, respectively. Next, we divide the sorted voxels into several groups of equal size $m$, resulting in a total of $\lceil L/m\rceil$ groups, where $\lceil \cdot \rceil$ donates ceiling function. Since the number of voxels varies across different input point clouds and is rarely an integer multiple of the size $m$, we employ an average overlap grouping method. This ensures that the number of overlapping voxels between any pair of adjacent groups is equal across the data. Furthermore, we utilize absolute position encoding to emphasize global characteristics within this module, defined as follows: 
$$
E_{i}=MLP(2\pi c_{i}) \eqno{(1)}
$$
where $MLP$ denotes a Multilayer Perceptron that projects the spatial coordinates $2\pi c_{i}$ from 3-dimensional space to a d-dimensional space consistent with the feature dimension. Here, $c_{i}$ represents the normalized position of voxel $i$, computed as:
$$
c_{i}=(x_{i}/X,y_{i}/Y,z_{i}/Z) \eqno{(2)}
$$
where $(X,Y,Z)$ represents the voxel range of the scene, and $(x_{i},y_{i},z_{i})$ represents the coordinates of voxel $i$.

\subsection{Grouped-Transformer Block}

\begin{figure}
    \centering
    \includegraphics[width=0.45\textwidth]{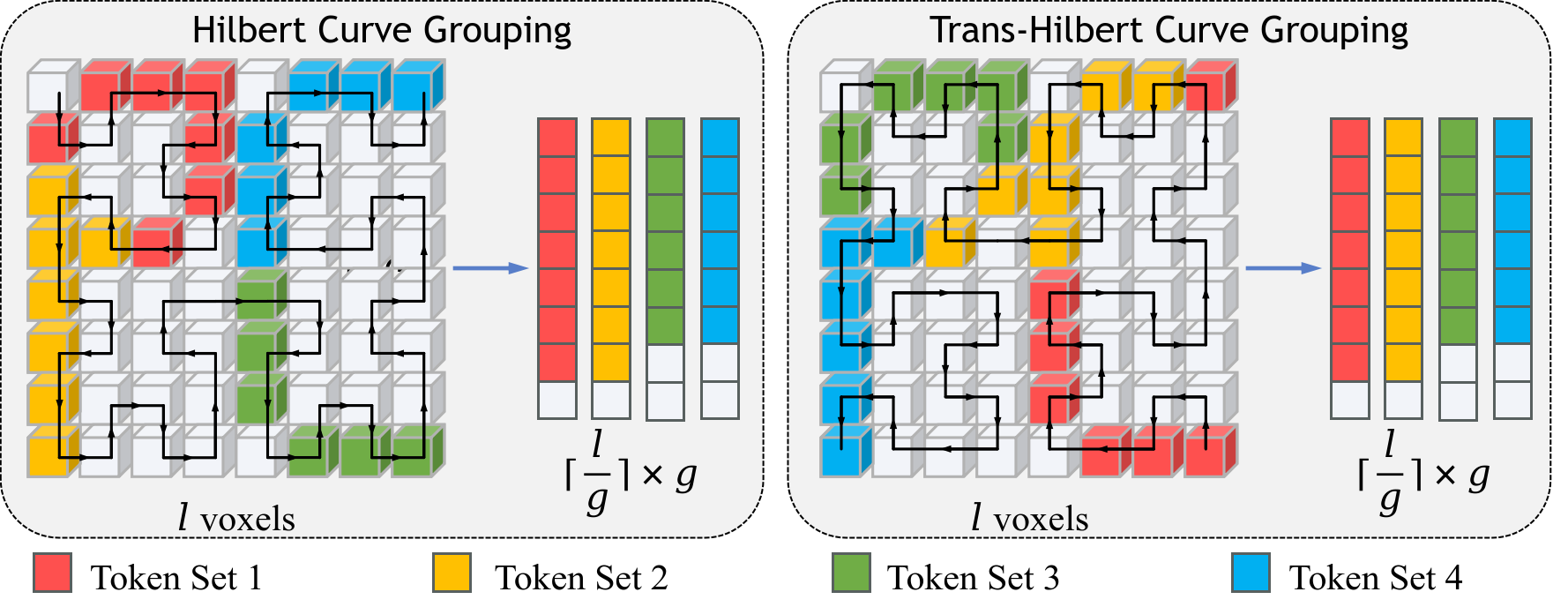}
    \caption{Illustration of the serialization scheme in Grouped-Transformer Block. All voxels are divided into fixed size windows. The non-empty voxels in the windows are grouped into constant size g in the order of Hilbert or Trans-Hilbert curve, and arranged from low to high along the Z-axis. Empty voxels are padded into the groups to ensure that all groups within a window have an equal number of non-empty voxels.}
    \label{figurelabel_transformer_grouping}
\end{figure}

Our Transformer module adopts a window-based grouping scheme to extract local features. As shown in figure \ref{figurelabel_transformer_grouping}, we first partition non-empty voxels into non-overlapping 3D windows of shape $(T_{x},T_{y},T_{z})$, where $T_{x}$, $T_{y}$, and $T_{z}$ denote the length, width, and height of the window along the $X$, $Y$, and $X$ axes, respectively. Then, we sort the voxels within these windows along the Hilbert curve and the Trans-Hilbert curve, respectively. Next, we divide these non-empty voxels into groups of constant size $g$. Consequently, assuming a window contains $l$ non-empty voxels, the window is divided into $\lceil l/g\rceil$ groups. The voxels are evenly distributed among these groups. Any deficiency in group size is padded with empty voxels, which are subsequently eliminated via masking during the attention operation.

We utilize an attention mechanism to extract local features. The standard attention mechanism is defined as:
$$
Attention(Q,K,V)=softmax(\frac{Q{K}^T}{\sqrt{d_{k}}})V \eqno{(3)}
$$
where, $Q(Query)$, $K(Key)$ and $V(Value)$ represent the input projection vectors. The attention mechanism calculates the similarity (attention score) between $Q$ and $K$, utilizing this score to perform a weighted summation on $V$ to obtain the final result.

To further enhance feature capture capabilities, we adopt a cross-attention mechanism. The core difference is that the $Query$ originates from one sequence, while the $Key$ and $Value$ originate from another. The cross-attention mechanism is defined as:

$$
CrossAttention_{H}(Q_{H},K_{HT},V_{HT})=$$$$softmax(\frac{Q_{H}{K_{HT}}^T}{\sqrt{d_{k}}})V_{HT} \eqno{(4)}
$$
$$
CrossAttention_{HT}(Q_{HT},K_{H},V_{H})=$$$$softmax(\frac{Q_{HT}{K_{H}}^T}{\sqrt{d_{k}}})V_{H} \eqno{(5)}
$$
where, $Q_{H}$, $K_{H}$ and $V_{H}$ represent vectors from the Hilbert curve sequence, while $Q_{HT}$, $K_{HT}$ and $V_{HT}$ represent vectors from the Trans-Hilbert curve sequence.

\begin{figure}
    \centering
    
    \includegraphics[width=0.45\textwidth]{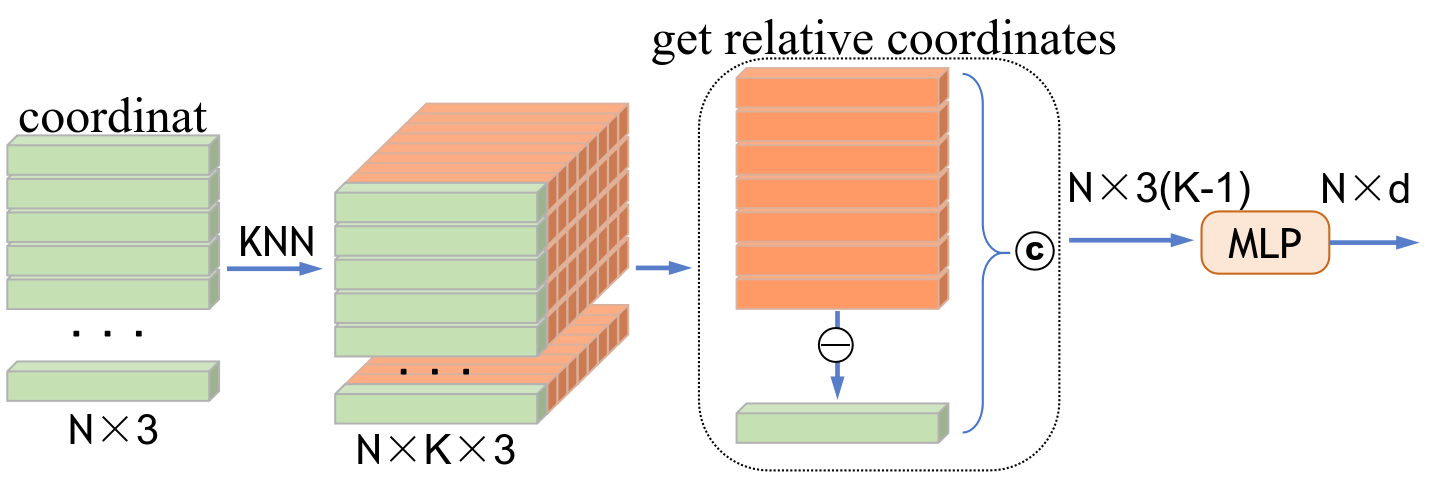}
    \caption{Illustration of our relative position encoding. Each voxel first computes the difference with its K-nearest neighboring voxels to obtain the relative position. Then we directly encode the relative position through a MLP and then fused with main features.}
    \label{figurelabel_relative_code}
\end{figure}

The local neighborhood contains key features of the point cloud data. The Grouped-Transformer block utilizes relative position encoding to capture local geometric characteristics. We observe that the coordinate difference between a voxel and its adjacent voxels can be interpreted as their relative position within the point cloud. As shown in Fig. \ref{figurelabel_relative_code}, we first obtain the K-Nearest Neighbors (KNN) for each voxel. We then calculate the relative positions of the voxels and their neighbors to obtain $K-1$ groups of neighboring relative coordinates. Finally, an $MLP$ is used to encode these relative coordinates.

\subsection{Voxel Generated Block}
\begin{figure}
    \centering
    \includegraphics[width=0.45\textwidth]{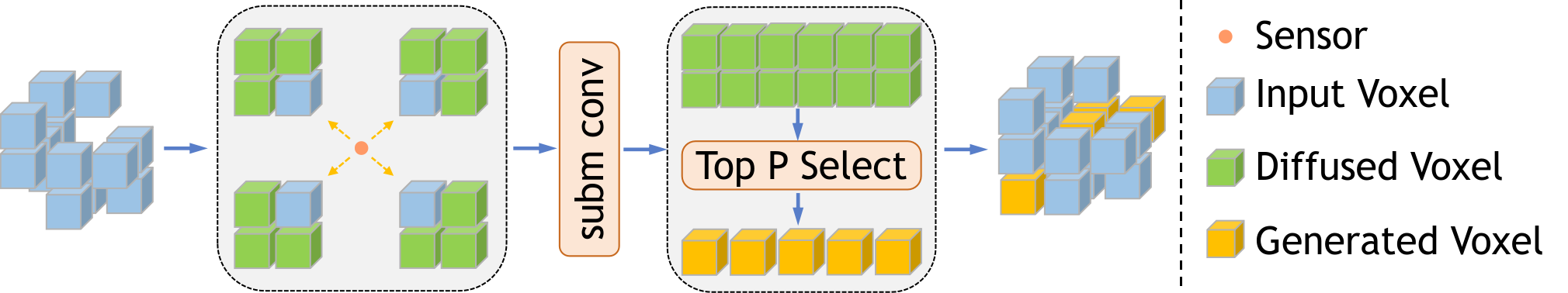}
    \caption{Illustrations of our voxel generation scheme. The voxels are diffused away from the sensor, and the top-P diffused voxels are selected as the generated voxel.}
    \label{figurelabel_Generation}
\end{figure}
Considering the challenges of feature representation in highly sparse point clouds, we propose a voxel generation strategy based on the imaging characteristics of LiDAR sensors.

Standard voxelization methods treat empty space as a nullity, failing to account for the physical imaging characteristics of LiDAR. We argue that the space immediately behind a detected point is not "empty" but "unknown," with a high probability of occupancy.

As shown in Fig. \ref{figurelabel_Generation}, we first determine the sensor position $(X_{s},Y_{s},Z_{s})$. Then, each voxel diffuses to generate three new voxels in directions moving away from the sensor. The diffused voxels are formulated as follows:
$$
F_{d}^{i}={F_{i[R_{Xi}, 0, 0]}}\oplus{F_{i[0, R_{Yi}, 0]}}\oplus{F_{i[R_{Xi}, R_{Yi}, 0]}} \eqno{(6)}
$$
$$
{R_{Xi}}=\frac{X_{i}-X_{s}}{|{X_{i}-X_{s}}|}, {R_{Yi}}=\frac{Y_{i}-Y_{s}}{|{Y_{i}-Y_{s}}|} \eqno{(7)}
$$
where $F_{i}$ denotes the feature of initial voxel $i(i=1,2,...,N)$ , and $F_{d}^{i}$ are the diffused voxels features of voxel $i$. The symbol $\oplus$ denotes the concatenation operation. $R_{Xi}$ and $R_{Yi}$ denote the diffusion direction signs of voxel $i$ along the X-axis and Y-axis, respectively. The term $F_{[x, y, z]}$ denotes the diffused voxel features with offsets of 1 along the axes, where the value is initialized equal to the input voxel featur $F_{i}$.

Subsequently, we extract 3D features using a 3D sparse convolution. Next, we select top-P voxels from the total number $L_{d}$ of diffused voxels, where $P=L_{d}\times r$ and $r$ is the ratio of generation. Finally, we concatenate these generated voxels with the voxels in the initial feature map to form the output of this block:
$$
{F_{ini}^{'}}\oplus{F_{d}^{'}}=subm({F_{ini}}\oplus{F_{d}}) \eqno{(8)}
$$
$$
{F_{G}}={F_{ini}^{'}}\oplus{top_{P}(F_{d}^{'})} \eqno{(9)}
$$
where $subm$ denotes submanifold sparse convolution, $top_{P}(F_{d}^{'})$ means selecting Top-P voxel features from $F_{d}^{'}$. $F_{ini}$ and $F_{d}$ are the initial voxels and diffused voxels features, and $F_{G}$ represents the final generated voxels.

\section{EXPERIMENTS}


\begin{table*}[h]
\centering
\caption{Comparison with Previous Methods on Kitti Validation set. The \textbf{bolded} and \underline{underlined} values are the best and second-best performance, respectively.}
\label{table_kitti}
\resizebox{0.98\linewidth}{!}
{\begin{tabular}{c|cccc|cccc|cccc|c}
\hline
\multicolumn{1}{c|}{\multirow{2}{*}{Method}} &\multicolumn{4}{c|}{Car} &\multicolumn{4}{c|}{Pedestrian} &\multicolumn{4}{c|}{Cyclist} &\multirow{2}{*}{mAP}\\
 &Easy &Moderate &Hard &Overall &Easy &Moderate &Hard &Overall &Easy &Moderate &Hard &Overall & \\
\hline
PointRCNN\cite{PointRCNN}	&85.94 	&75.76 	&68.32 &76.67	&49.43 	&41.78 	&38.63 &43.28	&73.93 	&59.60 	&53.59 	&62.37 &60.78\\
PointPillars\cite{PointPillars}	&79.05 	&74.99 	&68.30 &74.11	&52.08 	&43.53 	&41.49 &45.7	&75.78 	&59.07 	&52.92 	&62.59 &59.20\\
SECOND\cite{SECOND}	&\textbf{88.51} 	&\textbf{78.19}	&76.01 &\textbf{80.90}	&52.12	&46.08 	&42.06 &46.75	&78.48	&64.64	&60.57	&67.90 &65.18\\
CenterPoint\cite{CenterPoint}	&79.26	&69.21 	&64.80	&71.09 &32.96	&30.19 	&28.03 &30.39	&69.92 	&53.29 	&51.35 	&58.19 &53.22\\
DSVT\cite{DSVT}	&87.78	&77.21	&74.93	&79.97 &58.56	&52.12	&47.46	&52.71 &\textbf{82.50}	&\underline{65.08}	&\underline{62.81}	&\underline{70.13} &67.61\\
LION\cite{LION}	&\underline{88.12}	&77.84 	&\underline{76.51} 	&80.82 &\underline{62.11}	&\underline{55.82}	&\underline{50.48} &\underline{56.14}	&79.39	&61.37 	&57.93 	&66.23 &\underline{67.73}\\
\hline
\textbf{Ours}	&87.95	&\underline{78.0}	&\textbf{76.63}	&\underline{80.86}	&\textbf{62.60}	&\textbf{57.55}	&\textbf{51.61}	&\textbf{57.25}	&\underline{82.20}	&\textbf{67.74}	&\textbf{63.96}	&\textbf{71.30}	&\textbf{69.80}

\\
\hline
\end{tabular}}
\label{KITTI}
\end{table*}
\begin{table*}[h]
\centering
\caption{Comparison with Previous Methods on ONCE Validation set. The \textbf{bolded} and \underline{underlined} values are the best and second-best performance, respectively.}
\label{table_once}
\resizebox{0.98\linewidth}{!}
{\begin{tabular}{c|cccc|cccc|cccc|c}
\hline
\multicolumn{1}{c|}{\multirow{2}{*}{Method}} &\multicolumn{4}{c|}{Vehicle} &\multicolumn{4}{c|}{Pedestrian} &\multicolumn{4}{c|}{Cyclist} &\multirow{2}{*}{mAP}\\
 &overall &0-30m &30-50m &50m-inf &overall &0-30m &30-50m &50m-inf &overall &0-30m &30-50m &50m-inf & \\
\hline
PointRCNN\cite{PointRCNN} &52.1 &74.5 &40.9 &16.8 &4.3 &6.2 &2.4 &0.9 &29.8 &46.0 &20.9 &5.5 &28.7\\
PointPillars\cite{PointPillars} &68.6 &80.9 &62.1 &47.0 &17.6 &19.7 &15.2 &10.2 &46.8 &58.3 &40.3 &25.9 &44.3 \\
SECOND\cite{SECOND} &71.2 &84.0 &63.0 &47.3 &26.4 &29.3 &24.1 &18.1 &58.0 &70.0 &52.4 &34.6 &51.9\\
PV-RCNN\cite{PV-RCNN} &{77.8} &\textbf{89.4} &{72.6} &\textbf{58.6} &23.5 &25.6 &22.8 &17.3 &59.4 &71.7 &52.6 &36.2 &53.6\\
PointPainting\cite{PointPainting} &66.2 &80.3 &59.8 &42.3 &44.8 &52.6 &36.6 &22.5 &62.3 &73.6 &57.2 &40.4 &57.8\\
CenterPoint\cite{CenterPoint} &66.8 &80.1 &59.6 &43.4 &49.9 &56.2 &42.6 &\textbf{26.3} &63.5 &74.3 &57.9 &41.5 &60.1\\
DSVT\cite{DSVT} &76.5 &88.4 &\underline{72.7} &54.2 &47.4 &55.2 &38.9 &24.0 &64.2 &75.4 &59.9 &41.0 &62.7\\
LION\cite{LION} &\underline{78.2} &\underline{89.1} &{72.6} &{57.5} &\underline{53.2} &62.4 &\textbf{44.0} &{24.5} &\underline{68.5} &\underline{79.2} &\underline{63.2} &\underline{47.1} &\underline{66.6}\\


\hline
\textbf{Ours} &\textbf{78.3} &88.9 &\textbf{72.9} &\underline{58.0} &\textbf{53.3} &\textbf{62.6} &\underline{43.8} &\underline{25.9} &\textbf{68.7} &\textbf{79.8} &\textbf{64.0} &\underline{46.1}  &\textbf{66.8}\\

\hline
\end{tabular}}
\label{ONCE}
\end{table*}

\subsection{Datasets and Settings}

Our work is implemented on OpenPCDet\cite{openpcdet2020}. We train our models on 4 NVIDIA 4090D GPUs for 80 epochs with a batch size of 8 , and with the same ADAM optimizer as Centerpoint. For KITTI dataset, we set the feature channel as 64, set the point cloud range as (0\textasciitilde{~}70.4m, -40\textasciitilde{~}40m, -3\textasciitilde{~}1m) and voxel size as (0.2m, 0.2m, 0.125m). For ONCE dataset, we set the feature channel as 128, set the point cloud range as (-75.2\textasciitilde{~}75.2m, -75.2\textasciitilde{~}75.2m, -5\textasciitilde{~}3m) and voxel size as (0.4m, 0.4m, 0.25m). 

The number of MAMTRA blocks $N$ is set to 5. The group size $m$ of Mamba block are set to 4096, 2048, 1024, 512, 512, respectively. The window shape $(T_{x},T_{y},T_{z})$ of Transformer block is set to (64, 64, 32) and the group size $g$ is set to 90. The number $K$ of nearest neighbors in relative position encoding is 8. The ratio $r$ of voxel generation is 0.2.

The comparison includes widely adopted voxel-based methods (SECOND\cite{SECOND}, PointPillars\cite{PointPillars}), advanced two-stage frameworks (PV-RCNN\cite{PV-RCNN}, CenterPoint\cite{CenterPoint}). In particular, we compared with DSVT\cite{DSVT} and LION\cite{LION}, which are the typical algorithmic applications of Transformer and MAMBA in the field of 3D object detection. We adopt the same BEV backbone, detect head and optimization operation as SECOND\cite{SECOND}, and this is the same as LION. Since DSVT has not been tested on the ONCE dataset, we migrated its DSVT-voxel 3D backbone to SECOND and retrained for comparison. 

KITTI Dataset. The KITTI Dataset consists of 7481 training frames and 7518 test frames for 3D object detection. We split the 7481 training frames into 3712 frames for training set and 3769 frames for validation set, follow the dataset splitting protocol of OpenPCDet\cite{openpcdet2020}. The KITTI dataset is divided into three categories: car, pedestrian, and cyclist for three difficulty levels, i.e., Easy, Moderate, and Hard. Due to inconsistencies in the evaluation metrics used in the published papers, we retrained and experimented with all algorithms using the code and configurations released by OpenPCDet\cite{openpcdet2020}. During training, we set the road plane to False. The mean Average Precision (mAP) across 11 recall levels is used as the evaluation metric.

ONCE Dataset. The ONCE dataset consists of 15K annotated LiDAR scenes, partitioned into 5000, 3000, and 8000 frames for training, validation, and testing set, respectively. We evaluate the performance by mean Average Precision (mAP) for three objects’ classes(Vehicle, Pedestrian, Cyclist).

\subsection{Comparison With State-of-The-Arts}

\textbf{Performance on KITTI Dataset}

Table \ref{KITTI} benchmarks our proposed 3DTMDet against state-of-the-art 3D object detectors on the KITTI dataset. Our 3DTMDet achieves a new state-of-the-art mean Average Precision (mAP) of 69.80, outperforming both the Mamba-based LION (67.73) and the Transformer-based DSVT (67.61). The effectiveness of our hybrid strategy is most evident on geometrically complex, small objects where local detail is critical. In the Pedestrian and Cyclist categories, we achieve overall mAPs of 57.25 and 71.3, representing substantial improvements over LION (+1.11 and +5.07, respectively). Gains are particularly significant in the Hard difficulty level, where objects are highly occluded or distant; notably, our Cyclist Hard mAP of 63.96 surpasses LION by 6 points. These results confirm that the Grouped-Transformer block effectively preserves the fine-grained local geometries that pure Mamba architectures tend to smooth, while the Voxel Generated Block restores structural continuity for sparse targets. For the Car category, our method remains highly competitive (80.86 mAP), confirming that integrating local attention does not compromise the global receptive field provided by Mamba.

The effectiveness of our hybrid strategy—synergizing Mamba for remote interaction and Transformers for local geometric modeling—is most evident in the performance on smaller, geometrically complex objects. While pure Mamba-based architectures like LION excel at capturing global context, their serialization process often degrades local structural details. This limitation is effectively addressed by our Grouped-Transformer block.

\textbf{Performance on ONCE Dataset}

To further validate the generalization ability of 3DTMDet, we conducted comparisons on the KITTI dataset, as shown in Table \ref{ONCE} . In general, 3DTMDet attains an overall mAP of 66.8, achieving the best results for each category (78.3, 53.3, and 68.7 for Vehicle, Pedestrian, and Cyclist). It consistently outperforms DSVT (+4.1 mAP) and demonstrates a modest improvement over LION (+0.2 mAP). We attribute the smaller gain over LION on this dataset to the sparser LiDAR data (40-beam vs. 64-beam in KITTI), which inherently provides less local geometric structure for the Grouped-Transformer to exploit. Under such extreme sparsity, the model relies more on the global context capability of the Serialized-Mamba block, delivering performance on par with the LION baseline. This analysis confirms that our hybrid design is particularly advantageous in high-fidelity perception scenarios with sufficient beam density.

In summary, the experiments demonstrate that 3DTMDet is not merely a combination of modules, but a complementary architecture. By leveraging Mamba for efficient remote feature interaction and Transformers for high-fidelity local 3D geometric modeling, we achieve a robust balance that resolves the local-detail blindness of SSMs while mitigating the computational costs of global Transformer.

\subsection{Ablation Studies}

In this section, a set of ablative studies are conducted on ONCE validation split to investigate the key designs of our model. 

\begin{table}[h]
\centering
\caption{Ablation study on ONCE validation split for each component in 3DHMT Block. We removed or replaced the SSM or Transformer blocks in the 3DHMT Block to validate the effectiveness of the proposed method, SBM represents Serialized-Mamba Block, GTB represents Grouped-Transformer Block. We also combined two SOTA algorithms(LION, DSVT) for experimentation to demonstrate that our algorithm is not simply stacked.}
\label{table_ab_1}
\begin{tabularx}{0.5\textwidth}{XXXXXp{1.3cm}}
\hline
SSM Block &Trans-former Block &Vehicle &Pedes-trian &Cyclist &Overall\\
\hline
SBM &$\times$ &76.2 &48.7 &65.9 & 63.6\\
$\times$ &GTB &74.2 &48.2 &64.1 & 62.2\\
SBM &SpConv &77.6 &51.5 &68.8 &65.9\\
SpConv &GTB &76.3 &52.7 &67.1 &65.4\\
LION &DSVT &78.1 &53.1 &68.3 &66.5\\
\hline
SBM &GTB &78.3 &53.3 &68.7 &\textbf{66.8(Ours)}\\
\hline
\end{tabularx}
\end{table}

\begin{figure}
    \centering
    \includegraphics[width=0.2\textwidth]{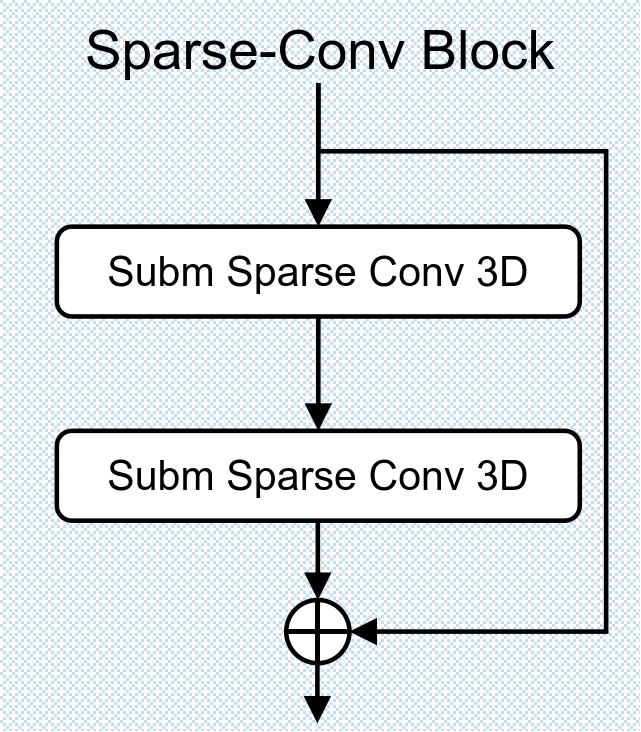}
    \caption{Illustration of the basic sparse-conv block. This module contains two 3D Subm Sparse Convolutional layers and a residual link. This work replaces the Transformer Blocks or Mamba Blocks with this module to verify the effectiveness of our proposed network structure.}
    \label{sparse-conv}
\end{figure}

Table \ref{table_ab_1} shows ablation studies on our major structures,  which are Serialized Mamba Block(SBM) and Grouped Transformer Block(GTB). The combination of these tow modules constitutes the main body of our framework. By removing these tow modules, we verified their necessity. Compared with the pure Mamba and Transformer architectures, our algorithm improves 3.2mAP and 4.6mAP, demonstrating the superiority of our hybrid strategy. In addition, unlike directly removing these two modules, we designed an experiment using basic sparse conv blocks to replace these two modules separately. This module includes two 3D Subm Sparse Convolutional layers and a residual link, as shown in Figure \ref{sparse-conv} In the specific operation, each of our SBM modules or GTB modules is replaced by two basic sparse conv blocks, while the other parts remain unchanged. As shown in the experiment, our algorithm improves 0.9mAP and 1.4mAP. Furthermore, in order to compare with the algorithm with the best performance, we combined the two existing best algorithms: based on the LION algorithm, we added a DSVT block after each LION block. The experiment achieved 66.5mAP, slightly 0.3mAP lower than our algorithm, which proves that our algorithm is not simply a stack of Mamba and Transformer, but an organic combination of the two.

\begin{table}[h]
\centering
\caption{Result of attention and position encoding type ablation experiments. We modified the Position Encoding strategy in the serialized Mamba Block (SMB) and Grouped Transformer Block (GTB). We also conducted experiments on the use of cross-attention or self-attention in the second Attention Layer of Transformer Block.}
\label{table_att}
\begin{threeparttable} 
\begin{tabularx}{0.48\textwidth}{XXXX}
\hline
Attention Layer &Position Encoding(SMB) &Position Encoding(GTB) &mAP(overall)\\
\hline
cross   &no         &no         & 65.4\\
cross   &absolute   &absolute   & 65.6\\
self    &absolute   &relative   & 66.4\\
\hline
cross   &absolute   &relative   & \textbf{66.8(Ours)}\\
\hline
\end{tabularx}
\end{threeparttable} %
\end{table}

Table\ref{table_att} shows ablation studies on the operations of our structures, in which, SMB stands for the Serialized-Mamba Block and GTB stands for the Grouped-Transformer Block. To validate the effectiveness of our position encoding operations, we compare different strategies. To be specific, we remove the Position Encoding block of Serialized Mamba Block and Grouped Transformer Block, performance decreases to 65.4mAP, indicating that the absence of Position Encoding can significantly degrade performance. We also replaced the Relative Position Encoding of the Grouped Transformer Block with Absolute Position Encoding. Experiments show that our Relative Position Encoding outperforms Absolute Position Encoding by 1.2mAP. Furthermore, we substitute the cross-attention with original self-attention. Our cross-attention performance improved 0.4mAP, which illustrates the benefits of performing cross-attention for 3D feature interaction.

\begin{table}[h]
\centering
\caption{Ablation experiments on ONCE validation split for voxel generation type. We selected different voxel generation strategies or without generating voxels to verify the effectiveness of our voxel enhancement method.}
\label{table_ab_VG}
\begin{tabularx}{0.38\textwidth}{lll}
\hline
Voxel Generation Method & &mAP(overall)\\
\hline
no Generation & & 65.6\\
Sparse-Conv & & 66.1\\
no Top P Select & & 66.3\\
\hline
Ours Generation & & \textbf{66.8 (Ours)}\\
\hline
\end{tabularx}
\end{table}

Table \ref{table_ab_VG} shows the ablation experiment on Voxel Generation Block. We observe that the design of Voxel Generation even brings 1.2 mAP performance improvement, which illustrates the benefits of voxel generation strategy for target feature enhancement. In addition, we conducted ablation experiments on the voxel generation strategy, using sparse conv for diffusion and eliminating the Top P Select step in our strategy. The experiment shows that our algorithm outperforms these two strategies by 0.7mAP and 0.5mAP, respectively. This indicates that selective diffusion is an effective means of enhancing the effectiveness of angle measurement.

\begin{figure}
	\centering   %
	\subfigure[Input voxels map] %
	{
		\begin{minipage}[b]{0.46\linewidth}
			\centering
			\includegraphics[width=1\linewidth]{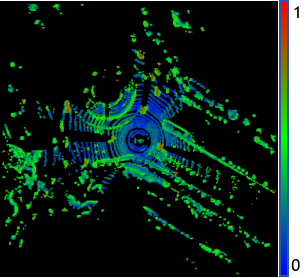}
		\end{minipage}
	}
	\subfigure[Diffused voxels map]
	{
		\begin{minipage}[b]{0.46\linewidth}
			\centering
			\includegraphics[width=1\linewidth]{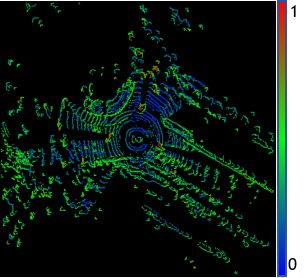}
		\end{minipage}
	}
    
	\subfigure[Generated voxels map]
	{
		\begin{minipage}[b]{0.46\linewidth}
			\centering
			\includegraphics[width=1\linewidth]{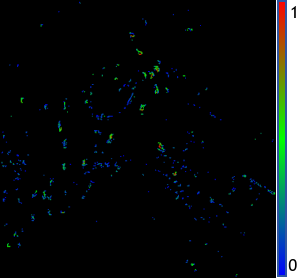}
		\end{minipage}
	}
    \subfigure[Generated voxels with ground truth boxes]
	{
		\begin{minipage}[b]{0.46\linewidth}
			\centering
			\includegraphics[width=1\linewidth]{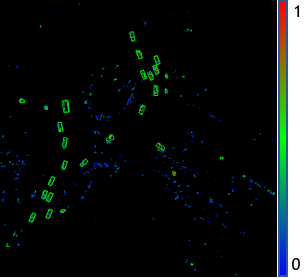}
		\end{minipage}
	}
	\caption{Heatmap of voxel generation block}

    \label{figurelabel_heatmapGeneration}
\end{figure}

In this section, we also present visualizations of the heatmap of the generated voxels in Voxel Generation block in Fig.\ref{figurelabel_heatmapGeneration}. In this figure, Fig.\ref{figurelabel_heatmapGeneration}(a) is the input voxel map, and Fig.\ref{figurelabel_heatmapGeneration}(b) is the unfiltered voxels map after diffusion. After screening out the voxeles with the highest score in Fig.(b), we obtain the Fig.\ref{figurelabel_heatmapGeneration}(c). Fig.\ref{figurelabel_heatmapGeneration}(d) is obtained by drawing the true value with a green box on Fig.\ref{figurelabel_heatmapGeneration}(c). As illustrated in the figure, points within the ground truth boxes achieve the highest score, verified the correctness of our screening strategy and further validating the effectiveness of the VG module we proposed. 

\begin{figure*}[htbp]
    \centering
    \vspace*{3mm} 
    \includegraphics[width=0.95\textwidth, keepaspectratio]{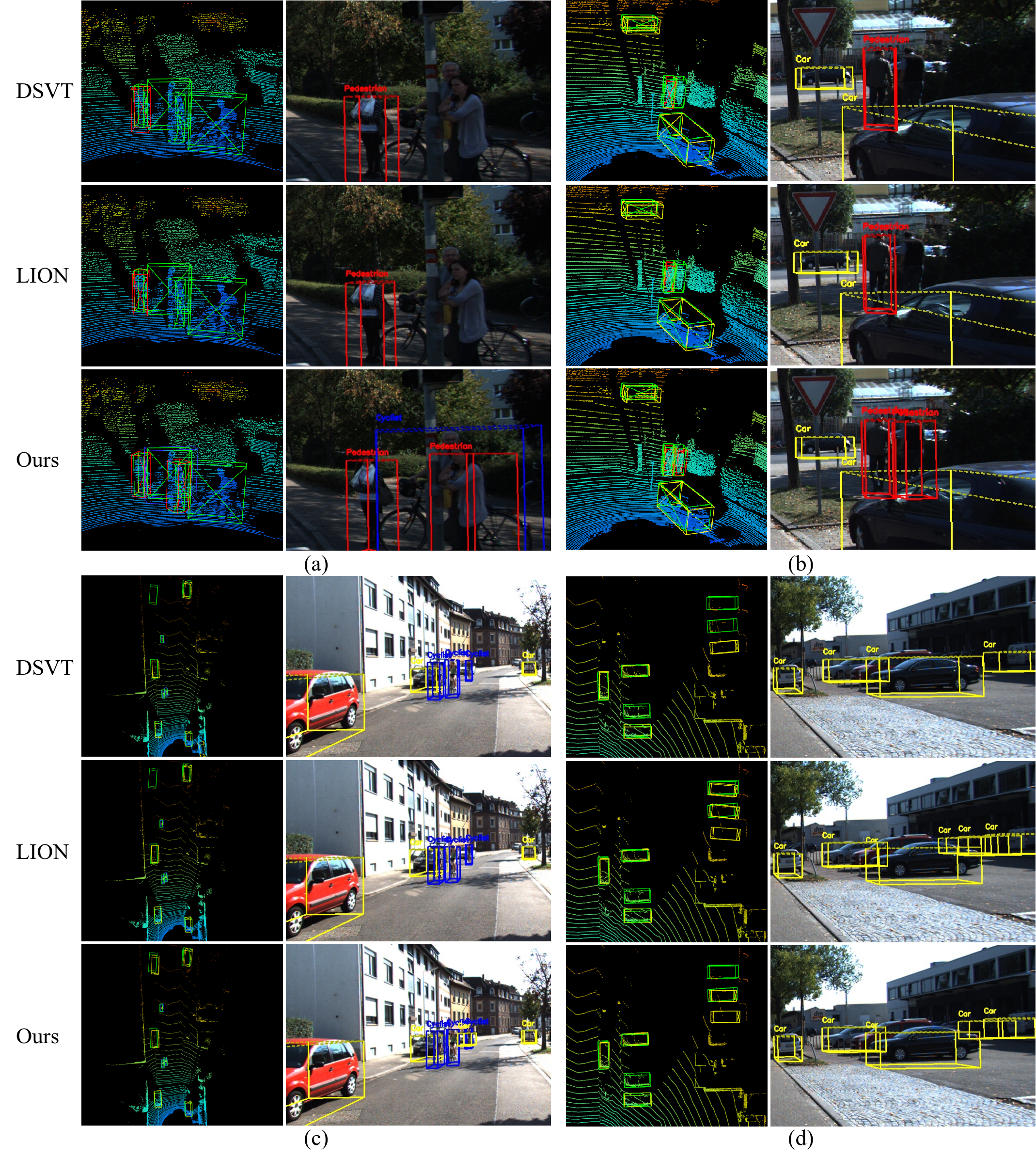}
    \caption{The Qualitative Results Visualization of DSVT, LION and Ours.}
    \label{QualitativeVisualization}
    
\end{figure*}

\subsection{Qualitative Results Studies}
In this section, we present detection results visualizations of our method and  in Fig.\ref{QualitativeVisualization}, and compared with LION and DSVT. These studies selected four valsets from the Kitti dataset, namely 000211, 001071, 002690, and 005430, with a threshold of 0.7. In the four sets of visualizations, the left column displays the detection results of point clouds and the label boxes are shown with green boxes, while the right column shows the effect of projecting the detection boxes onto visible light images.

In images Fig.\ref{QualitativeVisualization}(a) and Fig.\ref{QualitativeVisualization}(b), it can be seen that our algorithm exhibits stronger detection capabilities for Cyclinst and Pedestrian targets. Fig.\ref{QualitativeVisualization}(c) shows that our algorithm also maintains good detection performance for distant Car targets. This proves that for small objects such as pedestrians and cyclists, or for targets with long distances, the geometric information provided by point clouds is limited, and our strategy can better ensure detection performance. In Fig.\ref{QualitativeVisualization}(d), our algorithm, compared to the pure Mamba algorithm Lion, missed detecting Car targets, but still outperformed the pure Transformer algorithm DSVT.

\section{Conclusion}
This paper presents 3DTMDet, a dual-path synergy neural network that integrates Transformer and Mamba for point cloud 3D object detection. The core innovation is the 3DHMT block, which uses bidirectional serialized Mamba for efficient global modeling and cross-attention grouped Transformer for high-fidelity local extraction. A physics-inspired voxel generation module further alleviates point cloud sparsity.

3DTMDet is not a straightforward application of existing models, but a systematic neural architecture design that addresses fundamental conflicts between long-range efficiency and geometric precision in 3D representation learning. Experiments on KITTI and ONCE demonstrate its superiority over state-of-the-art Transformer and Mamba detectors.

In the future, we will extend this hybrid paradigm to multi-modal fusion, large-scale semantic segmentation, and real-time embedded perception systems.


\bibliographystyle{cas-model2-names}

\bibliography{cas-refs}



\end{document}